\documentclass[sn-vancouver]{sn-jnl}

\usepackage{graphicx}
\usepackage{subfigure}
\usepackage{cleveref}
\Crefname{figure}{Fig.}{Figs.}

\jyear{2021}%

\theoremstyle{thmstyleone}%
\theoremstyle{thmstyletwo}%

\theoremstyle{thmstylethree}%

\begin{document}

\title[Graph-Based Multi-Robot Path Finding and Planning]{Graph-Based Multi-Robot Path Finding and Planning}


\author*[1]{\fnm{Hang} \sur{Ma}}\email{hangma@sfu.ca}

\affil*[1]{\orgdiv{School of Computing Science}, \orgname{Simon Fraser University}, \orgaddress{\street{8888 University Drive}, \city{Burnaby}, \postcode{V5A 1B5}, \state{BC}, \country{Canada}}}


\abstract{
\noindent\textbf{Purpose of Review} 
Planning collision-free paths for multiple robots is important for real-world multi-robot systems and has been studied as an optimization problem on graphs, called Multi-Agent Path Finding (MAPF). This review surveys different categories of classic and state-of-the-art MAPF algorithms and different research attempts to tackle the challenges of generalizing MAPF techniques to real-world scenarios.
	
\noindent\textbf{Recent Findings} 
Solving MAPF problems optimally is computationally challenging. Recent advances have resulted in MAPF algorithms that can compute collision-free paths for hundreds of robots and thousands of navigation tasks in seconds of runtime. Many variants of MAPF have been formalized to adapt MAPF techniques to different real-world requirements, such as considerations of robot kinematics, online optimization for real-time systems, and the integration of task assignment and path planning.

\noindent\textbf{Summary}
Algorithmic techniques for MAPF problems have addressed important aspects of several multi-robot applications, including automated warehouse fulfillment and sortation, automated train scheduling, and navigation of non-holonomic robots and quadcopters. This showcases their potential for real-world applications of large-scale multi-robot systems.
}

\keywords{Multi-Agent Path Finding; Multi-Robot Path Planning; Multi-Robot Systems\\ \\This preprint has not undergone peer review (when applicable) or any post-submission improvements or corrections. The Version of Record of this article is published in Current Robotics Reports, and is available online at \url{https://doi.org/10.1007/s43154-022-00083-8}.}



\maketitle

\section{Introduction}\label{sec1}

In many real-world multi-robot systems, robots have to plan collision-free paths to different locations to execute different tasks. Today, thousands of warehouse robots already navigate fully autonomously to relocate inventory pods in automated fulfillment centers \cite{kiva,honig2019persistent} or deliver parcels in sortation centers \cite{KouAAAI20}. In the coming years, autonomous aircraft-towing vehicles will tow aircraft from the runways to the terminal gates (and vice versa) at airports. Other examples include autonomous intersection management \cite{dresner2008multiagent}, forklift robot fleets \cite{pecora2018loosely,salvado2018motion}, game characters in video games \cite{MaAIIDE17}, object-transportation robots \cite{rus1995moving}, patrolling robots \cite{agmon2011multiagent}, service robots \cite{ahmadi2006multi,khandelwal2017bwibots}, swarms of differential-drive robots and quadcopters \cite{HoenigIROS16,preiss2017downwash,honig2018trajectory}, robots in formations \cite{LiAAMAS20}, and other multi-robot systems \cite{MaIEEE17}. 

%
%


Solving the path planning problem optimally for multiple robots is computationally challenging, especially for a large number of robots. However, the above real-world applications require computing high-quality collision-free paths for a large number of robots in a short computation time since shorter paths result in higher throughput or lower operating costs (since fewer robots are required to achieve the same throughput) of the systems.

\section{Multi-Agent Path Finding (MAPF)}

Many recent works in the artificial intelligence, robotics, and operations research communities have modeled the path planning problem for multiple robots as a combinatorial optimization problem on graphs, called Multi-Agent Path Finding (MAPF) \cite{MaAIMATTERS17,SoCS2019Surv}. MAPF has also been studied under the name of Multi-Robot Path Planning on Graphs \cite{yu2016optimal}. A MAPF problem instance consists of a connected undirected graph and a set of robots. The vertices of the given graph correspond to locations and the edges correspond to connections between locations that the robots can move along. Each robot occupies one vertex at each discrete time step and is given a start vertex and a goal vertex. Between two consecutive time steps, each robot takes an action to either move to an adjacent vertex or wait at its current vertex. Two robots collide if they move to the same vertex or traverse the same edge in opposite directions at the same time. The problem of MAPF is to find collision-free paths for the robots from their start vertices to their goal vertices. The objective is to minimize either the makespan, defined as the maximum of the arrival times of all robots at their goal vertices, or the flowtime, defined as the sum of the arrival times of all robots at their goal vertices.

Finding a solution to any MAPF problem instance or deciding its unsolvability can be done in polynomial time \cite{YuR14}. However, it is NP-hard (namely, unlikely that a polynomial-time algorithm exists) to find a solution with the minimum makespan \cite{surynek2010optimization} or the minimum flowtime \cite{YuLav13AAAI} to a MAPF problem instance, even if the given graph is a planar graph \cite{Yu16RAL} or a 2D 4-neighbor grid \cite{banfi2017intractability}. In addition, it is NP-hard to compute an approximate solution within any constant factor less than 4/3 to a MAPF problem instance \cite{MaAAAI16}.

On one hand, recent advances in MAPF solving have resulted in powerful MAPF algorithms that can compute collision-free paths for a large number of robots in a short runtime, despite the complexity of solving MAPF optimally. These advances have resulted in a number of achievements, including a MAPF software \cite{LiICAPS21} that recently won the Flatland Challenge
\cite{Laurent21},
a train-scheduling competition at NeurIPS 2020 (one of the top machine learning conferences). The MAPF solver has been demonstrated to be capable of computing high-quality paths (namely with small makespan or flowtime) for up to 3,000 robots in minutes of runtime on a simulator. On the other hand, there are key challenges \cite{MaWOMPF16} that must be addressed in order to apply MAPF algorithms to real-world applications of multi-robot systems, which requires techniques beyond MAPF solving. The following sections of this review survey latest advances that enhance MAPF solving and extensions to MAPF solving that tackle the research challenges in generalizing it to real-world scenarios.

\section{MAPF Algorithms}

Recent MAPF algorithms can be categorized into reduction-based, rule-based, and search-based algorithms. In the following, we survey their methodologies and highlight their properties in terms of completeness (complete for all MAPF problem instances, complete for MAPF problem instances on graphs with special properties, or incomplete) and optimality (optimal, bounded-suboptimal, or suboptimal with respect to different objectives). A MAPF algorithm is complete for a class of MAPF problem instances if it guarantees to return a solution for any solvable MAPF problem instance in the class or correctly decide that the given MAPF problem instance in the class is unsolvable in finite time.

\paragraph{}\noindent\textbf{Reduction-Based MAPF Algorithms} Reduction-based MAPF algorithms reduce MAPF to other well-studied combinatorial problems, such as Boolean Satisfiability \cite{surynek2012towards}, Integer Linear Programming \cite{YuLav16TOR}, and Answer Set Programming \cite{erdem2013general,GomezHB20}. They are complete for all MAPF problem instances. They solve MAPF with the makespan objective optimally but can be modified to solve MAPF with other objectives optimally \cite{DBLP:conf/ecai/SurynekFSB16,YuLav16TOR,GomezHB20}, bounded-suboptimally (with the guarantee that the resulting solution is within a user-provided suboptimality factor from the optimal solution) \cite{SurynekFSB17}, and suboptimally \cite{Surynek15,WangAAMAS19,han2019integer}. They perform well for MAPF problem instances on small-size graphs with densely placed robots. For example, a state-of-the-art Integer Linear Programming based MAPF solver can compute a solution with the minimum makespan for a MAPF problem instance on a $24\times18$ 2D 4-neighbor grid with 60 robots in less than 15 seconds of runtime \cite{han2019integer}.

\paragraph{}\noindent\textbf{Rule-Based MAPF Algorithms} Rule-based algorithms solve MAPF using a set of primitive operations that specify the actions of the robots in different situations. They often guarantee completeness for only a restricted class of MAPF problem instances. Rule-based algorithms are often very efficient by simply following the predefined primitive operations but provide no guarantee on the solution quality (optimality). Push and Swap \cite{PushAndSwap} and its extension \cite{PLPushAndSwap} can compute a solution for 100 agents in less than 10 seconds of runtime but provide no completeness guarantee theoretically. One of their descendants, Push and Rotate \cite{PushAndRotate}, is complete for MAPF problem instances on graphs with at least two vertices that are unoccupied by robots. TASS \cite{KhorshidHS11} is complete for MAPF problem instances on ``solvable'' trees based on prior work on solving multi-robot motion planning on trees \cite{masehian2009solvability}. BIBOX \cite{Surynek09} is complete for MAPF problem instances on bi-connected graphs with at least two vertices unoccupied by robots. Its descendant \cite{botea2018solving} works for strongly bi-connected directed graphs with at least two vertices unoccupied by robots. SAG \cite{yu2020average} is complete for MAPF problem instances on grid-like ``well-connected'' graphs, runs in polynomial time, and provides a constant-factor approximation guarantee for minimizing the makespan on such graphs.

\paragraph{}\noindent\textbf{Search-Based MAPF Algorithms} Search-based MAPF algorithms \cite{SoCS2017Surv} solve MAPF with heuristic search techniques. The main computational challenge of optimally solving MAPF with a search algorithm is that the number of possible states of a MAPF problem instance grows exponentially in the number of robots.
\begin{itemize}
	\item A*-based MAPF algorithms \cite{ODA,EPEJAIR,MStar} plan paths with joint states but try to reduce the size of the state space they need to explore. They are complete for all MAPF problem instances and can be used for either makespan minimization or flowtime minimization.
	\item Decoupled MAPF algorithms \cite{WHCA,WHCA06,bnaya2014conflict} plan paths for robots one at a time according to a predefined or a dynamic total ordering on the robots. Path planning for each robot uses an A* search in vertex and time dimensions that treats already planned paths of other robots as moving obstacles. Decoupled MAPF algorithms are often efficient but provide no optimality or even completeness guarantee. PIBT \cite{OkumuraMDT19} develops a scheme to decide a partial ordering on the robots dynamically but only guarantees that all robots reach their goal vertices at least once (but possibly not at the same time) in finite time on biconnected graphs. MAPF-LNS \cite{li2021anytime} is another recent decoupled MAPF algorithm that uses Large Neighborhood Search \cite{shaw1998using}, a local search algorithm, to improve a suboptimal MAPF solution by repeatedly replanning paths for a subset of robots. It is one of the core elements of the MAPF software \cite{LiICAPS21} that won the Flatland Challenge. Its descendant MAPF-LNS2 \cite{LiAAAI22} uses Large Neighborhood Search to improve a MAPF plan (paths of all robots) with collisions by repeatedly replanning paths for a subset of robots to reduce the number of collisions, until a collision-free MAPF plan (a MAPF solution) is obtained.
	\item Hierarchical MAPF algorithms plan paths for robots individually on the low level and dynamically couple the resulting single-robot paths with a tree search on the high level. They are complete for all MAPF problem instances. Increasing Cost Tree Search \cite{DBLP:journals/ai/SharonSGF13} minimizes the flowtime. It performs a best-first tree search of all combinations of the arrival times of robots on the high level and checks whether collision-free paths exist for a combination of arrival times on the low level. Conflict-Based Search (CBS) \cite{DBLP:journals/ai/SharonSFS15} is arguably the most popular optimal MAPF algorithm. It minimizes either the makespan or the flowtime. CBS first finds individually time-optimal paths for all robots (ignoring collisions). On the high level, it then performs a best-first search on a binary constraint tree. Each branching resolves one collision in the computed paths by imposing constraints on individual robots that forbid them from occupying a vertex or traversing an edge at a given time step. On the low level, CBS uses an A* search in vertex and time dimensions to replan for a robot that obeys the constraints. Many improvements to CBS have been proposed in recent years: Meta-Agent CBS \cite{DBLP:journals/ai/SharonSFS15} dynamically groups multiple robots into a meta-agent on the high level and uses an A* search to plan paths for these robots with their joint states on the low level. ICBS \cite{ICBS} always first resolves collisions that result in child search nodes whose costs are larger than that of the current node, thus affording the high-level search of CBS pruning opportunities. CBSH \cite{FelnerICAPS18} and its improvement \cite{LiIJCAI19} use an admissible heuristic to improve the high-level best-first search of CBS. Disjoint-Splitting CBS \cite{LiICAPS19} expands each node in a way such that any solution is admitted by the subtree under only one but not both of its child nodes, thus reducing duplicate search effort of the high-level search of CBS. IDCBS \cite{boyarski2021iterative} replaces the high-level best-first search of CBS with iterative-deepening depth-first searches. Symmetry-Breaking CBS \cite{LiAAAI19b,LiICAPS20,LiAIJ21} and Mutex-Propagation CBS \cite{ZhangICAPS20} add multiple constraints to a child node at a time to break symmetry in the high-level search of CBS. The best Symmetry-Breaking CBS variant has empirically been shown to compute optimal solutions for MAPF problem instances on a $256\times257$ 2D 4-neighbor grid with 100 robots in seconds of runtime \cite{LiAIJ21}. ECBS \cite{ECBS} and its improvements \cite{cohen2016improved,LiAAAI21a} perform a bounded-suboptimal search on the constraint tree, making CBS bounded-suboptimal. Recent research \cite{CohenIJCAI18} has also developed an anytime version of the bounded-suboptimal search on the constraint tree for CBS. Another line of recent research also uses machine learning to learn a good branching policy for the high-level search of CBS for both the optimal \cite{HuangKD21} and bounded-suboptimal \cite{HuangDK21} settings.
	\item Hybrid MAPF algorithms combine several of the above search-based MAPF techniques or combine search-based MAPF techniques with reduction-based or rule-based MAPF techniques. SMT-CBS \cite{Surynek19} replicates the high-level search of CBS with Satisfiability Modulo Theories, which is then solved by a Boolean Satisfiability solver, and minimizes either the makespan or the flowtime. Lazy CBS \cite{gange2019lazy} replaces the high-level search of CBS with a Constraint Programming solver and minimizes the flowtime. BCP \cite{lam2019branch,lam2020new} combines Branch-and-Cut-and-Price techniques for Mixed Integer Programming with symmetry-breaking techniques for MAPF and minimizes the flowtime. Priority-Based Search \cite{MaAAAI19a} is a recent hierarchically decoupled that performs a depth-first search on a binary priority tree to explore all possible orderings on the robots. It is complete for only ``well-formed'' MAPF problem instances and has empirically been shown to compute close-to-optimal solutions for MAPF problem instances on a $481\times530$ 2D 4-neighbor grid with 600 robots in half a minute of runtime. Some algorithms combine both primitive operations (rule-based MAPF techniques) and search. MAPP~\cite{WangB11} explore different ways of combining paths of individual robots and is complete for MAPF problem instances on ``slidable'' graphs \cite{WangB11}. There is also a MAPF algorithm that uses a combination of A* searches on a graph abstraction, primitive operations, and reductions to Constraint Satisfaction Problems~\cite{ryan2010constraint}.
\end{itemize}

Recent research \cite{kaduri2020algorithm,ren2021mapfast} has used machine learning to select a MAPF algorithm among multiple candidate MAPF algorithms for a given MAPF problem instance. 

\section{MAPF Extensions and Related Problems}

Recent studies have also generalized the standard definition of MAPF to different real-world scenarios.

\paragraph{}\noindent\textbf{MAPF with Deadlines}
MAPF with Deadlines \cite{MaAAMAS18,MaIJCAI18} aims to maximize the number of robots that reach their goal vertices within a given deadline. Its applications include robots that need to evacuate before a disaster and robots that need to finish tasks before a deadline.

\paragraph{}\noindent\textbf{MAPF with Delay Probabilities and Robust MAPF}
MAPF with Delay Probabilities (MAPF-DP) \cite{MaAAAI17} generalizes MAPF to the case where the uncertainty of robot motion has to be considered during planning to ensure a collision-free execution of the plan. In MAPF-DP, the uncertainty of each robot is characterized by a given delay probability with which the robot stays in its current vertex whenever it intends to traverse an outgoing edge of its current vertex. The problem of MAPF-DP is to find a plan that consists of a path for each robot and a plan-execution policy that controls with GO or STOP commands how each robot proceeds along its path such that no collisions occur during plan execution. MAPF-DP has also been studied under the name MAPF with Uncertainty \cite{wagner2017path}, where the paths are planned in the belief space of the robots and the execution of the resulting plan is not guaranteed to be collision-free. $K$-Robust MAPF \cite{AtzmonSFWBZ18} extends CBS to enforce $K$ time steps for which a vertex must be unoccupied after it has been occupied by a robot during planning, which reduces the possibility of collisions during plan execution without using plan-execution policies. Recent research \cite{ChenAAAI21b} has generalized Symmetry-Breaking CBS \cite{LiAIJ21} to the $K$-Robust MAPF setting. Probabilistic Robust MAPF \cite{atzmon2020probabilistic} bounds the probability that any collision occurs during plan execution.

\paragraph{}\noindent\textbf{MAPF with Continuous Time or Kinematic Constraints}
MAPF with Continuous Time \cite{andreychuk2022multi} extends CBS to planning paths on weighted graphs where the edge weights characterize the nonuniform traversal times of the edges. Other research \cite{bartak2018scheduling,walker2018extended} has also studied MAPF on weighted graphs. MAPF for Large Agents \cite{LiAAAI19a} allows a robot to occupy more than one vertex at one time step according to its given shape and volume. Two robots collide if both of them occupy some vertex at the same time step. The resulting CBS-based algorithm has been applied to planning collision-free trajectories for quadcopters that take into account their ellipsoid shapes and downwash effects. MAPF-POST \cite{HoenigICAPS16,HoenigIJCAI17} is a polynomial-time algorithm that post-processes a MAPF solution to create a plan-execution schedule that works on non-holonomic robots, takes their kinematic constraints, such as the maximum and minimum translational and rotational velocity limits, into account, and provides a guaranteed safety distance between them, which avoids time-intensive replanning in many cases.

\paragraph{}\noindent\textbf{Reinforcement Learning for Distributed MAPF}
PRIMAL \cite{sartoretti2019primal} and its descendant \cite{damani2021primal} model MAPF as a multi-agent reinforcement learning task, where all robots follow the same learned single-agent policy to decide their actions at each time step based on their local observations. Recent research \cite{li2020graph,MaICRA21,li2021message} uses a graph neural network to allow robots to communicate and also precomputed shortest path distances \cite{MaICRA21} or an online shortest path computation \cite{wang2020mobile,damani2021primal} to assist training. We note that classic optimization-based collision-avoidance approaches \cite{DBLP:conf/isrr/BergGLM09,guy2009clearpath,zhou2017fast} use a similar distributed MAPF setting but can be applied to robots moving in continuous space.

\paragraph{}\noindent\textbf{MAPF with Target Assignment}
Anonymous MAPF, also known as Permutation-Invariant MAPF or Unlabeled MAPF, does not assume predefined goal vertices for the robots and aims to find a one-to-one mapping from the given goal vertices to the robots and collision-free paths for the robots to their assigned goal vertices. Minimizing the makespan for Anonymous MAPF is polynomial-time solvable using a max-flow algorithm \cite{YuLav13STAR}. CBS-TA \cite{honig2018conflict} searches all possible assignments of goal vertices to robots and minimizes the flowtime for Anonymous MAPF. Combined Target Assignment and Path Finding (TAPF) \cite{MaAAMAS16} partitions the robots into teams where the problem of each team is an Anonymous MAPF problem. CBM \cite{MaAAMAS16} combines the high-level search of CBS and a low-level min-cost max-flow algorithm to minimize the makespan for TAPF. Recent research \cite{GTAPF,henkel2019optimal} has studied generalized TAPF problems where each robot needs to get assigned multiple goal vertices. MG-MAPF \cite{Surynek21} assumes that each robot is preassigned multiple unordered goal vertices and aims to compute collision-free paths for the robots to visit all goal vertices. MG-TAPF \cite{ZhongICRA22} aims to find a one-to-one mapping from the given tasks that each consists of a sequence of ordered goal vertices and collision-free paths for the robots that visit the goal vertices of their assigned tasks in the specified order.

\paragraph{}\noindent\textbf{Online MAPF and Multi-Agent Pickup and Delivery}
Online MAPF \cite{vsvancara2019online,MaICAPS21} assumes that each robot is assigned a new goal vertex by a black box once it reaches its current goal vertex. Recent research has conducted a theoretical study \cite{MaICAPS21} on the competitiveness of online MAPF algorithms (namely the performance gap between online and optimal offline MAPF algorithms). RHCR \cite{LiAAAI21b} generalizes (offline) MAPF algorithms to online MAPF by repeatedly replanning paths for the robots. One version of RHCR that uses Priority-Based Search \cite{MaAAAI19a} has been shown to compute paths for 1,000 robots on a $37\times77$ 4-neighbor grid in less than half a minute of runtime. Multi-Agent Pickup and Delivery (MAPD) \cite{MaAAMAS17} is a combined multi-robot task-allocation and path-planning problem. MAPD has first been studied in an online setting where robots have to constantly get assigned a stream of incoming tasks that are added to the system at unknown release times and plan collision-free paths to the pickup and delivery vertices of the tasks. Online MAPD algorithms \cite{MaAAMAS17} repeatedly apply task-assignment and MAPF algorithms to (re-)assign tasks to and (re-)plan paths for robots whenever a new task arrives or a robot becomes available for executing tasks. Recent research \cite{MaAAAI19b} has developed an online MAPD algorithm that considers kinematic constraints of robots directly during planning and shown experimentally that the algorithm can compute solutions for MAPD problem instances with 250 robots and 2,000 tasks within a total runtime of ten seconds. Offline MAPD \cite{LiuAAMAS19} considers tasks that are known a priori. Recent research \cite{KouAAAI20} has also considered an online TAPF/MAPD variant that aims to minimize the idle time of sorting stations---that is, when there are no warehouse robots servicing the sorting stations---in an automated sortation center. The resulting algorithm has been shown to compute solutions for 350 robots within two seconds of runtime on an industrial robot simulator.

\section{Conclusions}

Planning collision-free paths for multiple robots is a fundamental building block for many real-world applications of multi-robot systems. It has been studied as a graph-optimization problem under the name of MAPF by researchers from the artificial intelligence, robotics, and operations research communities. As the efficiency of MAPF algorithms improves, they will become increasingly viable for real-time path-planning operations of multi-robot systems. Current research has also addressed several challenges to adapt MAPF techniques to the requirements of real-world multi-robot systems. Future research directions include developing deeper theoretical understandings for MAPF variants such as Distributed MAPF and MAPD and combining MAPF techniques with considerations of complex real-world settings such as general temporal constraints and dependencies of tasks and high-order dynamic constraints of robots. Readers are referred to the MAPF information page
 \cite{MAPFinfo}
for a listing of MAPF researchers and links to their publications and software, tutorials, a mailing list, and other resources.

\paragraph{}\noindent\textbf{Acknowledgment} 
This work was supported by the Natural Sciences and Engineering Research Council (NSERC) of Canada under grant number RGPIN2020-06540.

\section*{Declarations}

\noindent\textbf{Conflict of Interest} 
The author declares no conflict of interest with the article.

\noindent\textbf{Human and Animal Rights} 
This article does not contain any studies with human or animal subjects performed by any
of the authors.

\paragraph{}
Papers of particular interest, published recently, have been highlighted as:

Of importance: \cite{YuLav13AAAI} \cite{LiAIJ21} \cite{MaAAAI19a} \cite{HoenigICAPS16} \cite{sartoretti2019primal} \cite{MaAAMAS17}

Of major importance: \cite{SoCS2019Surv} \cite{DBLP:journals/ai/SharonSFS15} 

\bibliography{references}


\end{document}